\documentclass[letterpaper]{article}
\usepackage{proceed2e}
\usepackage[margin=1in]{geometry}
\usepackage{graphicx}
\usepackage{amsmath}
\usepackage{amssymb}
\usepackage{bm}
\usepackage{algorithm2e}
\usepackage{xcolor}

\usepackage[round]{natbib}
\DeclareMathOperator{\cov}{cov}

 \usepackage{times}

\title{Online Constrained Model-based Reinforcement Learning}

\author{ {\bf Benjamin van Niekerk} \\
School of Computer Science\\
University of the Witwatersrand\\
South Africa \\
\And
{\bf Andreas Damianou\thanks{Work done while this author was at the University of Sheffield.}}  \\
Amazon.com \\
Cambridge, UK \\
\And
{\bf Benjamin Rosman}   \\
Council for Scientific and Industrial Research, \\
and School of Computer Science\\
University of the Witwatersrand\\
South Africa\\
}

\begin{document}

\maketitle

\begin{abstract}
Applying reinforcement learning to robotic systems poses a number of challenging problems. A key requirement is the ability to handle continuous state and action spaces while remaining within a limited time and resource budget. Additionally, for safe operation, the system must make robust decisions under hard constraints. To address these challenges, we propose a model based approach that combines Gaussian Process regression and Receding Horizon Control. Using sparse spectrum Gaussian Processes, we extend previous work by updating the dynamics model incrementally from a stream of sensory data. This results in an agent that can learn and plan in real-time under non-linear constraints. We test our approach on a cart pole swing-up environment and demonstrate the benefits of online learning on an autonomous racing task. The environment's dynamics are learned from limited training data and can be reused in new task instances without retraining.
\end{abstract}

\section{INTRODUCTION}

Reinforcement learning has become an important approach to the planning and control of autonomous agents in complex environments. However, recent interest in reinforcement learning is yet to be reflected in robotics applications; possibly due to their specific challenges.

In robotic systems, reinforcement learning methods must deal with continuous, and potentially high-dimensional, state and control spaces. For example, the dynamics of autonomous vehicles are most naturally described in terms of continuous variables like position, velocity, orientation and steering angle. Data efficiency is also critical since collecting experience, or evaluating control policies, with a robot in the loop can be costly and time consuming. Poorly modeled dynamics, due to a slow learning rate, may result in unstable trajectories and dangerous collisions. The system also needs to handle physical constraints --- from joint and actuator limitations to the boundaries defined by obstacles. Finally, in fast-paced applications rapid decisions, requiring real-time planning, must be made. 

In this paper we propose a method which tackles all of the above challenges simultaneously. Our method achieves this through the marriage of two key components: 
\begin{enumerate}
\item a learned dynamics model based on sparse spectrum Gaussian processes (GPs), and
\item a planner based on receding horizon control (RHC), and the structure exploiting interior point solver FORCES \citep{FORCESPro}.
\end{enumerate}
Sparse spectrum GPs allow us to update the learned model online and the RHC framework provides the ability plan in real-time while naturally handling constraints. This means that we can learn and plan online in constrained environments where data collection is expensive. We believe that these are important features for real world applications of reinforcement learning, particularly in safety critical systems. Our proposed method will be referred to as Gaussian Process-Receding Horizon Control (GP-RHC hereafter).

The remainder of the paper is structured as follows. In section \ref{sec:literature} we provide an overview of related approaches in model-based reinforcement learning. The receding horizon control framework is presented in section \ref{sec:rhc}. This is followed, in section \ref{sec:model}, by a discussion on the application of Gaussian process regression to model learning. In section \ref{sec:experiments} we apply GP-RHC to a cart pole swing-up environment and a challenging autonomous racing task. The results demonstrate that (a) models can be learned quickly from limited data, (b) complex non-linear constraints can be handled in real-time, and (c) online updates improve the rate of learning and result in more consistent performance.

\section{RELATED WORK}
\label{sec:literature}

GP-RHC falls into the class of model-based reinforcement learning methods. These are generally the methods of choice in robotics primarily due to their impressive data efficiency \citep{kober2013reinforcement}. Model-based approaches can be broadly classified according to (a) how the dynamics model is learned and, (b) the choice of planner.

PILCO \citep{deisenroth2011pilco}, for example, combines policy search for planning with a Gaussian process model of the dynamics. The policy search relies on analytic gradients of closed form solutions to the long term expected cost. This requires the cost function and policy to take specific functional forms, making it difficult or impossible to incorporate general constraints. In a similar vein, \cite{kim2004autonomous} use policy search with locally weighted linear regression to learn a controller for helicopter flight.

As an alternative, trajectory optimization based on differential dynamic programming is often used for planning. Methods such as PDDP \citep{pan2014probabilistic} and AGP-iLQR \citep{boedecker2014approximate} make use of this idea by combining dynamics models learned by locally weighted projection regression with either an iterative linear quadratic regulator or Gaussian as the planner. These planners can take simple box constraints into account but cannot handle general non-linear constraints.

Our work is most closely related to the RL-RCO method \citep{andersson2015model} which leverages sparse Gaussian processes for learning the dynamics and trajectory optimization based sequential quadratic programming. We improve upon RL-RCO by proposing an approach which allows the dynamics model to be updated online as the agent interacts with the environment. Furthermore, in contrast to the work of \citet{andersson2015model}, we present results that highlight the ability to handle non-linear constraints. 

We show how these enhancements improve data efficiency, learning rate and constraint handling, rendering GP-RHC overall more applicable to realistic scenarios.

\section{RECEDING HORIZON CONTROL}
\label{sec:rhc}

In this paper, we consider an agent operating in a continuous environment described by a set of differential equations, $\dot{\mathbf{x}}  = \mathbf{f}(\mathbf{x},\mathbf{u})$, where $\mathbf{x} \in \mathbb{R}^n$ represents the agent's state and $\mathbf{u} \in \mathbb{R}^m$ the control signal. The objective of the agent is to select a control $\mathbf{u}$ to minimize a cost function \eqref{eq:opt-problem-cost}, while conforming to the system dynamics, and additional constraints \eqref{eq:box-contraint-control}-\eqref{eq:non-linear-constraint}.

We address this overall problem using receding horizon control where the idea is to plan over a finite horizon by iteratively solving an optimization problem. Given a measurement of the current state $\hat{\mathbf{x}}_0$, a trajectory through the state-control space is calculated, and the first step of the control signal is applied to the system. The horizon is then shifted forward and the process repeats. Formally, at each time step the agent must minimize the cost function
\begin{subequations}
\label{eq:opt-problem}
\begin{equation}
\label{eq:opt-problem-cost}
J(\mathbf{x}_0) = h(\mathbf{x}(t_0 + T)) + \int_{t_0}^{t_0+T} \mathcal{L}(\mathbf{x}(t), \mathbf{u}(t))dt,
\end{equation}
where $t_0$ is the current time, $T$ denotes the length of the planning horizon, $\mathcal{L}$ is an intermediate cost function and $h$ is a terminal cost function. In order to ensure that the trajectory is feasible, the optimization is subject to the constraints:
\begin{align}
& \mathbf{x}(t_0) = \hat{\mathbf{x}}_0, \nonumber \\
& \dot{\mathbf{x}}(t) = \mathbf{f}(\mathbf{x}(t), \mathbf{u}(t)), \nonumber \\
& \underline{\mathbf{u}} \leq \mathbf{u}(t) \leq \overline{\mathbf{u}}, \label{eq:box-contraint-control} \\
& \underline{\mathbf{x}} \leq \mathbf{x}(t) \leq \overline{\mathbf{x}}, \label{eq:box-constraint-state} \\
& \mathbf{g}(\mathbf{x}(t), \mathbf{u}(t)) \leq \mathbf{0}, \label{eq:non-linear-constraint} \\
& \text{for all $t$ in $[t_0, t_0 + T]$ \nonumber }.
\end{align}
\end{subequations}
Inequalities \eqref{eq:box-contraint-control} and \eqref{eq:box-constraint-state} specify box constraints on the control and state spaces (described by upper and lower bounds), and can represent operational limits on actuators or joints. On the other hand, \eqref{eq:non-linear-constraint} can express non-linear, non-convex constraints through the function $\mathbf{g}$. This can be used, for example, to define road boundaries or obstacles in an autonomous driving task. 

The receding horizon formulation differs from other reinforcement learning approaches in a few important ways. First, instead of explicitly maintaining a representation of a policy or value function, the control is recalculated (over the shifting horizon) at each time step. This has the advantage of not requiring a representation for the entire problem \emph{a priori}, but incurs the additional computational burden of repeatedly replanning. Second, hard constraints can be explicitly specified and handled through \eqref{eq:non-linear-constraint}. This allows agents to safely learn by avoiding dangerous regions of the state and control spaces.

Without some simplifying assumptions we cannot solve problem \eqref{eq:opt-problem} directly. However, efficient approximate solutions can be found by following the ``first discretize, then optimize'' approach described in the sections below.

\subsection{DIRECT MULTIPLE SHOOTING}

Using direct multiple shooting \citep{bock1984multiple}, problem  \eqref{eq:opt-problem} can be transformed into a structured non-linear program (NLP). First, the time horizon $[t_0, t_0 + T]$ is partitioned into $N$ equal subintervals $[t_k, t_{k+1}]$ for $k=0,\ldots,N-1$. Then, taking a piecewise constant approximation of the control signal over each interval, a sequence of initial value problems, 
\begin{equation}
\label{eq:ivp}
\dot{\mathbf{x}}(t) = \mathbf{f}(\mathbf{x}(t), \mathbf{u}_k),  \ \mathbf{x}(t_k) = \mathbf{x}_k, \ t \in [t_k, t_{k+1}],
\end{equation}
can be set up for the state trajectory. Here, the variables $\mathbf{x}_k$ have been added as initial values. Each of these problems can then be integrated to obtain a discretized trajectory. However, in order to enforce continuity over the planning horizon, matching constrains,
\begin{align*}
\mathbf{x}_{k+1} &= \mathbf{F}_k(\mathbf{x}_k, \mathbf{u}_k), & k = 0,\ldots,N-1,
\end{align*}
are placed on $\mathbf{x}_k$ at the boundary of each subinterval. The functions $\mathbf{F}_k$ represent the solutions to the initial value problems \eqref{eq:ivp} at time $t_{k+1}$. 

Finally, the cost function is discretized over each time interval, resulting in the following NLP:
\begin{subequations}
\label{eq:nlp}
\begin{equation}
\label{eq:nlp-cost}
\min_{\mathbf{x},\mathbf{u}} \quad h(\mathbf{x}_N) +  \sum_{k=0}^{N-1}\mathcal{L}(\mathbf{x}_k, \mathbf{u}_k),
\end{equation}
subject to
\begin{align}
& \mathbf{x}_0 = \hat{\mathbf{x}}_0,  \label{eq:nlp-initial-constraint} \\
& \mathbf{x}_{k+1} = \mathbf{F}_{k}(\mathbf{x}_k, \mathbf{u}_k), & k&=0,\ldots,N-1 \label{eq:nlp-dynamics-model} \\
& \underline{\mathbf{u}} \leq \mathbf{u}_k \leq \overline{\mathbf{u}}, & k&=0,\ldots,N-1 \\
& \underline{\mathbf{x}} \leq \mathbf{x}_k \leq \overline{\mathbf{x}},  & k&=1,\ldots,N \\
& \mathbf{g}(\mathbf{x}_k, \mathbf{u}_k) \leq \mathbf{0}.  & k&=1,\ldots,N-1 \label{eq:nlp-nonlinear-constraint}
\end{align}
\end{subequations}
In principle, any non-linear programming method can be used to solve the above problem \citep{nocedal2006numerical}. However, the computational burden of solving an NLP at each time step is a severe limitation in real-time applications. To address this issue, there has been much interest in efficient optimization for receding horizon control \citep{domahidi2012efficient, vukov2013auto} . A particularly successful approach has been the combination of sequential quadratic programming (SQP) with structure exploiting stage-wise solvers \citep{kouzoupis2015towards}.

\subsection{SEQUENTIAL QUADRATIC PROGRAMMING}

In the SQP framework, the NLP \eqref{eq:nlp} is linearized about a given nominal trajectory $\mathbf{w}=[\mathbf{x}_0, \mathbf{u}_0, \ldots, \mathbf{x}_N]$. The trajectory can then be improved according to the update \citep{nocedal2006numerical},
\begin{equation}
\label{eq:trajectory-update}
\mathbf{w}^+ = \mathbf{w} + \alpha\Delta \mathbf{w}^*,
\end{equation}
where $\alpha$ is the step size and $\Delta \mathbf{w}^*$ denotes the solution to the following quadratic program:
\begin{subequations}
\label{eq:qp}
\begin{equation}
\label{eq:qp-cost}
\begin{split}
\min_{\Delta\mathbf{x}, \Delta\mathbf{u}} & \frac{1}{2} \sum_{k=0}^{N-1} \begin{bmatrix}
\Delta \mathbf{u}_k \\
\Delta \mathbf{x}_k \\
1
\end{bmatrix}^\intercal 
\begin{bmatrix}
\mathbf{R}_k & \mathbf{S}_k & \mathbf{r}_k \\
\mathbf{S}^\intercal_k & \mathbf{Q}_k & \mathbf{q}_k \\
\mathbf{r}^\intercal_k & \mathbf{q}^\intercal_k & \rho_k
\end{bmatrix}
\begin{bmatrix}
\Delta \mathbf{u}_k \\
\Delta \mathbf{x}_k \\
1
\end{bmatrix}  \\
& + \frac{1}{2} \Delta \mathbf{x}^\intercal_N \mathbf{Q}_N \Delta \mathbf{x}_N + \Delta \mathbf{x}_N \mathbf{q}_N + \rho_N,
\end{split}
\end{equation}
subject to the constraints
\begin{align}
& \Delta \mathbf{x}_0 = \hat{\mathbf{x}}_0 - \mathbf{x}_0, \\
& \Delta \mathbf{x}_{k+1} = \mathbf{A}_k \Delta \mathbf{x}_k + \mathbf{B}_k \Delta \mathbf{u}_k + \mathbf{a}_k,  \\
& \underline{\mathbf{u}} - \mathbf{u}_k \leq \Delta \mathbf{u}_k \leq \overline{\mathbf{u}} - \mathbf{u}_k,  \\
& \underline{\mathbf{x}} - \mathbf{x}_k \leq \Delta \mathbf{x}_k \leq \overline{\mathbf{x}} - \mathbf{x}_k,  \\
& \mathbf{G}_k \Delta \mathbf{x}_k + \mathbf{H}_k \Delta \mathbf{u}_k \leq \mathbf{g}_k.
\end{align}
\end{subequations}
The objective function \eqref{eq:qp-cost} is a quadratic approximation of the discretized cost function \eqref{eq:nlp-cost} and the constraints correspond to linearizations of \eqref{eq:nlp-initial-constraint} to \eqref{eq:nlp-nonlinear-constraint}. 

Provided that $\mathbf{Q}_k$ is positive semidefinite and $\mathbf{R}_k$ is strictly positive definite, the trajectory can be shown to converge to a local optimum by iteratively linearizing and optimizing \citep{nocedal2006numerical}. This is guaranteed for least squares cost functions popular in tracking and stabilization tasks. When a more general cost function is desired a positive definite approximation to the Hessian can be obtained using BFGS updates. 

In this paper we use FORCES \citep{FORCESPro} as the stage-wise solver for the quadratic program \eqref{eq:qp}. FORCES is an interior-point method tailored for problems arising in RHC. In particular, the block diagonal structure of the Hessian \eqref{eq:qp-cost} and the fact that the states are only directly coupled to the previous time step are exploited to give linear computational complexity in the planning horizon.

\begin{figure}[h]
\includegraphics[width=0.5\textwidth]{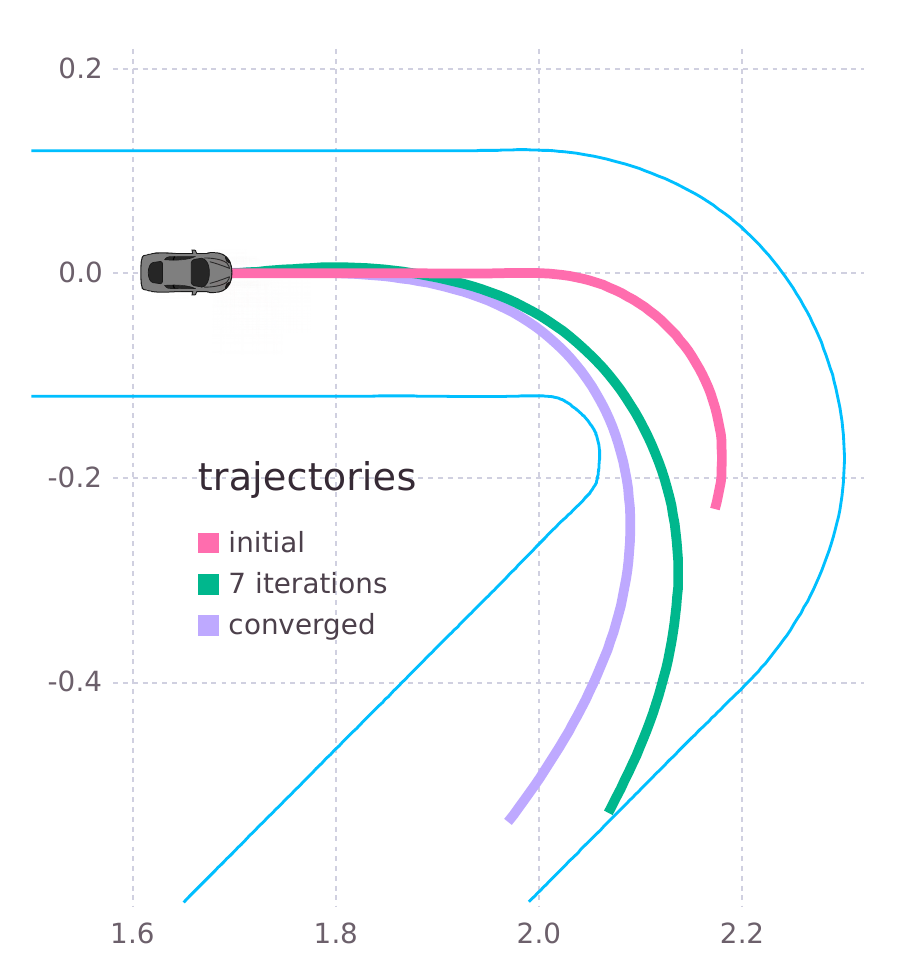}
\caption{An example of planned trajectories in an autonomous racing task. The agent's objective is to maximize its progress along the racing track over the planning horizon. Note how early termination yields a solution trajectory that is still near-optimal over a short horizon. }
\label{fig:cartrajectories}
\end{figure}

It is often unnecessary to iterate to convergence before a reasonable improvement is found. In real-time settings this is important because we need to maintain a balance between efficiency and accuracy. An example of this trade-off can be seen in figure \ref{fig:cartrajectories}. Given an initial trajectory, the car must maximize its progress along the race track.

The complete algorithm is presented in algorithm \ref{alg:method}. In the next section we discuss how sparse online Gaussian process can be used to learn the non-linear dynamics model \eqref{eq:nlp-dynamics-model}.

\begin{algorithm}
 \KwData{number of features $D$, initial training data}
 \ForEach{episode}{
 	train GPs on accumulated data (minimize \eqref{eq:loglikelihood})\;
 	linearise dynamics and constraints about $\mathbf{x}_0$\;
 	solve \eqref{eq:qp} until convergence\;
 	\While{not terminal}{
  		shift previous trajectory\;
  		\For{$i = 1$ to max iterations}{
  			linearise about current trajectory\;
  			solve \eqref{eq:qp} to get step direction\;
  			update trajectory \eqref{eq:trajectory-update}\;
  		}
  		apply control to the system\;
  		update dynamics model \eqref{eq:incremental-updates}\;
 	}
 }
\caption{The complete GP-RHC algorithm}
\label{alg:method}
\end{algorithm}

\section{DYNAMICS MODEL LEARNING}
\label{sec:model}

Typically, trajectory optimization based on receding horizon control relies on an analytic model of the system's dynamics. However, in the setting of reinforcement learning the dynamics are unknown and must be learned by interacting with the environment. In order to effectively plan over the horizon, an accurate model of the system's dynamics needs to be learned quickly.

Gaussian process (GP) regression is a non-parametric, Bayesian approach to model learning that has demonstrated impressive data-efficiency on both simulated and real robotic systems \citep{deisenroth2011pilco}. Unfortunately, GPs do not scale well to large data sets, limiting their applicability in practice. There are, however, a number of approximation schemes that significantly reduce computational costs \citep{quinonero2005unifying}. In this paper, we use a sparse spectrum approximation of the kernel function \citep{lazaro2010sparse}. Sparse spectrum GPs were chosen because (a) online data can be incorporated through incremental updates \citep{gijsberts2013real}, and (b) the learned model can be efficiently linearized to form the sequential quadratic programs.

\subsection{GAUSSIAN PROCESS REGRESSION}

Given an input set of $N$ state-control pairs $\tilde{\mathbf{x}} = (\mathbf{x}, \mathbf{u}) \in \mathbb{R}^{n+m}$ and the resulting (possibly noisy) state transitions $\Delta \mathbf{x} \in \mathbb{R}^n$, GP regression can be used to learn a model of the underlying dynamics. Following \citet{deisenroth2011pilco}, we train conditionally independent GPs on each component of the state transition vector, so for the remainder of this section we will represent target data as $y \in \mathbb{R}$.

Formally, a Gaussian process is a collection of random variables, any finite number of which have a joint Gaussian distribution \citep{rasmussen2006gaussian}. To completely specify a GP we need to choose a mean function $m(\tilde{\mathbf{x}})$ and a covariance function $k(\tilde{\mathbf{x}},\tilde{\mathbf{x}}')$, parametrized by a set of hyperparameters. A common, but flexible choice for the covariance function is the squared exponential kernel
\[
k(\tilde{\mathbf{x}}, \tilde{\mathbf{x}}') = \sigma^2 \exp \left(-\frac{1}{2}\sum_{i=1}^{n+m} \left(\frac{\tilde{x}_i - \tilde{x}'_i}{l_i}\right)^2\right),
\]
where the signal variance $\sigma^2$ and characteristic length scales $l_i$ constitute the kernel's set of hyperparameters. When modelling noisy targets, an additive noise term with variance $\sigma^2_n$ is included in the covariance function. 

Since we model the relative rather than absolute state transitions, we choose a zero mean prior. This means that in the absence of data, the state is expected to remain unchanged regardless of the control input.

After defining the input matrix $\mathbf{X} = [\tilde{\mathbf{x}}_1, \ldots, \tilde{\mathbf{x}}_N]$ and the corresponding target vector $\mathbf{y} = [y_1, \ldots, y_N]^\intercal$, the posterior predictive distribution for a set of test points, $\mathbf{X}_*$, is a multivariate Gaussian with mean
\begin{equation}
\label{eq:pred-mean}
\mathbb{E}[\mathbf{y}_*|\mathbf{X}_*,\mathbf{X}, \mathbf{y}] = \mathbf{K}(\mathbf{X}_*, \mathbf{X})\mathbf{Q}^{-1} \mathbf{y},
\end{equation}
and covariance
\begin{equation}
\label{eq:pred-cov}
\mathbf{K}(\mathbf{X}_*,\mathbf{X}_*) - \mathbf{K}(\mathbf{X}_*, \mathbf{X})\mathbf{Q}^{-1}\mathbf{K}(\mathbf{X},\mathbf{X}_*),
\end{equation}
where $\mathbf{Q} = \mathbf{K}(\mathbf{X},\mathbf{X}) + \sigma^2_n \mathbf{I}$, and $\mathbf{K}(\mathbf{X}, \mathbf{X})$ denotes the matrix of covariances evaluated at all pairs of input points. Computing the predictive mean and covariance are $\mathcal{O}(N)$ and $\mathcal{O}(N^2)$ respectively.

A common approach for learning the hyperparameters and the noise variance $\sigma^2_n$ is to maximize the marginal likelihood \citep{rasmussen2006gaussian}. The negative log marginal likelihood, given by
\begin{equation}
\label{eq:log-likelihood}
\mathcal{L} = \frac{1}{2} \log |\mathbf{Q}| + \frac{1}{2} \mathbf{y}^\intercal \mathbf{Q}^{-1} \mathbf{y} + \frac{n}{2}\log(2\pi),
\end{equation}
can be minimized using gradient based optimizers. Since $\mathbf{Q}$ needs to be inverted each time the log marginal likelihood is evaluated, which is an $\mathcal{O}(N^3)$ operation in general,  hyperparameter inference represents the main bottleneck for GP regression.

\subsection{SPARSE SPECTRUM APPROXIMATION}
In this section, we assume that the covariance function is stationary, i.e. that $k(\tilde{\mathbf{x}}, \tilde{\mathbf{x}}')$ is a function of $\mathbf{r}=\tilde{\mathbf{x}} - \tilde{\mathbf{x}}'$. In this case, Bochner's theorem \citep{rudin2011fourier} states that $k(\mathbf{r})$ can be represented as the Fourier transform,
\begin{equation}
\label{eq:fourier-trans}
k(\mathbf{r}) = \int_{\mathbb{R}^{n+m}} e^{i \bm{\omega}^\intercal \mathbf{r}}d\mu(\bm{\omega}),
\end{equation}
of a positive finite measure $\mu$. If $\mu(\bm{\omega})$ has a density, then it is called the power spectrum $S(\bm{\omega})$ of the covariance function and, by the Wiener-Khintchine theorem \citep{carlson2009communication}, $S(\bm{\omega})$ is the Fourier dual of $k(\mathbf{r})$. In particular, this means that $S$ is proportional to some probability measure $p$ over $\mathbb{R}^{n+m}$, and so equation \eqref{eq:fourier-trans} can be rewritten as an expectation
\[
k(\mathbf{r}) = \alpha^2 \mathbb{E}_p[e^{i\bm{\omega}^\intercal \mathbf{r}}],
\]
where $\alpha$ is the constant of proportionality. 

To approximate the expectation, we draw $D$ sample frequencies $\bm{\omega}_1,\ldots,\bm{\omega}_D$ from $p$ and take averages. Since the power spectrum is symmetric about zero, we also include $\bm{\omega}_{-j} = - \bm{\omega}_j$ for each sample frequency, in order to guarantee that $k(\mathbf{r})$ is real valued for all $\mathbf{r}$. This results in the sparse spectrum approximation \citep{lazaro2010sparse} of the covariance function:
\begin{equation}
\label{eq:ssgp-approx}
k(\tilde{\mathbf{x}},\tilde{\mathbf{x}}') \approx \frac{\alpha^2}{2D} \sum_{j=-D}^D e^{i\bm{\omega}_j^\intercal(\tilde{\mathbf{x}}-\tilde{\mathbf{x}}')}.
\end{equation}
In the particular case of the squared exponential kernel $\alpha^2=\sigma^2$, and the frequencies are drawn from the normal distribution $\mathcal{N}(\mathbf{0}, \mathbf{\Lambda}^{-1})$, where $\bm{\Lambda}=\operatorname{diag}([l^2_1,\ldots,l^2_{n+m}])$.

For convenience, we define the feature mapping $\bm{\phi} : \mathbb{R}^{n+m} \to \mathbb{R}^{2D}$ by,
\[
\begin{split}
\bm{\phi}(\tilde{\mathbf{x}}) = \frac{\alpha}{\sqrt{D}}
[
\ \cos(\bm{\omega}_1^\intercal & \tilde{\mathbf{x}}),\ \sin(\bm{\omega}_1^\intercal\tilde{\mathbf{x}}),\ \cdots ,\\ & \cos(\bm{\omega}_D^\intercal\tilde{\mathbf{x}}),\ \sin(\bm{\omega}_D^\intercal\tilde{\mathbf{x}})
\ ]^\intercal.
\end{split}
\]
In order to make use of the approximation \eqref{eq:ssgp-approx}, the matrix inversion lemma is applied to equations \eqref{eq:pred-mean} and \eqref{eq:pred-cov}, giving
\begin{align*}
\mathbb{E}[\mathbf{y}_*] & =\bm{\phi}(\mathbf{X}_*)^\intercal \mathbf{A}^{-1} \bm{\phi}(\mathbf{X})^\intercal \mathbf{y}, \\
\cov[\mathbf{y}_*] & = \sigma^2_n \bm{\phi}(\mathbf{X}_*)^\intercal \mathbf{A}^{-1} \bm{\phi}(\mathbf{X}_*),
\end{align*}
where $\mathbf{A} = \bm{\phi}(\mathbf{X})^\intercal \bm{\phi}(\mathbf{X}) + \sigma^2_n \mathbf{I}$, and $\bm{\phi}(\mathbf{X})$ is the matrix obtained by applying $\bm{\phi}$ to each column of $\mathbf{X}$. Instead of inverting the $N\times N$ matrix $\mathbf{Q}$, we now only require the inverse of the $2D\times2D$ matrix $\mathbf{A}$, which constitutes a significant saving in computational cost if $D \ll N$. Importantly, the size of $\mathbf{A}$ is independent of the number of training points, which makes it amenable to incremental updates \citep{gijsberts2013real}.

Applying the same idea to the negative log likelihood \eqref{eq:log-likelihood} gives the expression
\begin{equation}
\label{eq:loglikelihood}
\begin{split}
\mathcal{L} = \frac{1}{2} \log| & \mathbf{A}| - \frac{D}{2} \log \sigma^2_n + \frac{n}{2} \log(2 \pi \sigma^2_n) \\ + \frac{1}{2 \sigma^2_n} & \left(\mathbf{y}^\intercal \mathbf{y} - \mathbf{y}^\intercal \bm{\phi}(\mathbf{X}_*) \mathbf{A}^{-1} \bm{\phi}(\mathbf{X}_*)^\intercal \mathbf{y}\right).
\end{split} 
\end{equation}
Again, the smaller size of $\mathbf{A}$ results in reduced computational complexity for hyperparameter inference. In particular, each step of the gradient based optimization is $\mathcal{O}(ND^2)$.

\subsection{INCREMENTAL UPDATES} \label{sec:incremental-updates}
To incrementally handle a stream of data, the matrix $\mathbf{A}$ and the vector $\mathbf{b} = \mathbf{\Phi}^\intercal \mathbf{y}$ need to be updated in real time. Given a new sample, $(\tilde{\mathbf{x}}, y)$, the updates are computed according to the rules:
\begin{equation}
\label{eq:incremental-updates}
\mathbf{A} \leftarrow \mathbf{A} + \phi(\tilde{\mathbf{x}})\phi(\tilde{\mathbf{x}})^\intercal \quad \text{and} \quad \mathbf{b} \leftarrow \mathbf{b} + \phi(\tilde{\mathbf{x}})y.
\end{equation}
Since $\mathbf{A}$ remains positive semidefinite after each update, we do not need to store it explicitly. Instead, we can keep track of its upper triangular Cholesky factor. This allows us to make use of fast, numerically stable rank-1 Cholesky updates \citep{gijsberts2013real}.

\section{EXPERIMENTS}
\label{sec:experiments}

In the following sections we evaluate the performance of GP-RHC on a cart-pole swing up task and an autonomous racing scenario. In all the experiments GP-RHC was able to run in real-time with appropriate choices of the sparsity $D$ and planning horizon $N$. The choice of $D$ essentially involves a trade-off between model accuracy and the computational costs of both the learning and the prediction routines. Empirically, we found that a value between 20 and 100 allowed us to learn a sufficiently accurate dynamics model while remaining within the real-time constraints. Similarly, the choice of $N$ involves a balance between computational costs and the quality of the controller. Ideally, one would choose a large value of $N$ so that the controller can optimally react to future changes in dynamics or constraints.

\subsection{CARTPOLE SWINGUP}

Cartpole experiments are a common benchmark in both reinforcement learning and control theory \citep{kober2013reinforcement}. The basic set-up consists of a cart with an attached pendulum running along a track. In the swing-up task, the pendulum is initially pointing downwards and the objective is to apply horizontal forces to the cart in order to swing the pendulum up and balance it above the cart in the center of the track. This is a relatively difficult control problem as the dynamics are fairly non-linear.  Additionally, a long planning horizon is required because the cart must be pushed back and forth in order to develop enough momentum to swing the pendulum up.

The state of the system, $\mathbf{x} = [x, v, \theta, \omega]$, is described by the position of the cart, the velocity of the cart, the angle of the pendulum and its angular velocity. A horizontal force $u$ in the range of $-10$N to $10$N can be applied to the cart at a sampling rate of $0.025$s. The cost function is given by a least squares objective penalizing the distance from the set point $ [0, 0, \pi, 0]$.

\begin{figure}[h]
\includegraphics[width=0.5\textwidth]{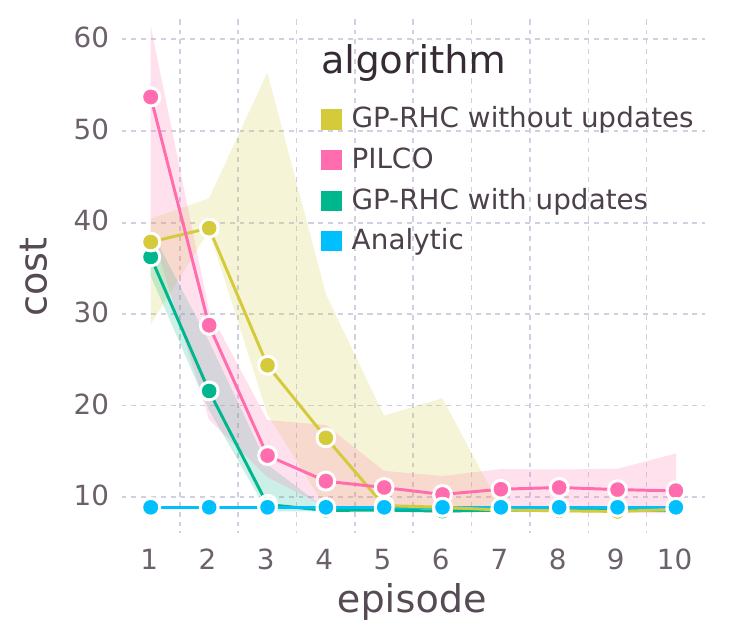}
\caption{Median cost of the unconstrained cartpole swing-up task (using the saturated cost function). The lower and upper bounds of the confidence envelope represents the first and third quantiles respectively. The results were aggregated over 10 runs.}
\label{fig:cartpole-comparison}
\end{figure}

In the first experiment we compare GP-RHC against PILCO \citep{deisenroth2011pilco} and the ground-truth analytical model. To initialize each trial, 80 data points were collected from an episode with random control inputs. For the sparse spectrum approximation 50 sample frequencies were drawn and a planning horizon of 50 time steps was used. After each episode the GP models were retrained on all the preceding data using 10 random restarts to avoid poor local minima. PILCO can potentially have problems with least squares costs \citep{deisenroth2010efficient} so we used (the preferred) saturating cost function and post-processed the results. Figure~\ref{fig:cartpole-comparison} shows that GP-RHC is competitive with PILCO both in terms of sample efficiency and overall performance.

In the second experiment the length of the track is limited; constraining the cart's position to between $-2$m and $2$m. Since PILCO does not handle constraints we just compare GP-RHC to the analytical model. The GP models were initialized with 20 points of training data collected from an episode with random inputs.

\begin{figure}[h]
\includegraphics[width=0.5\textwidth]{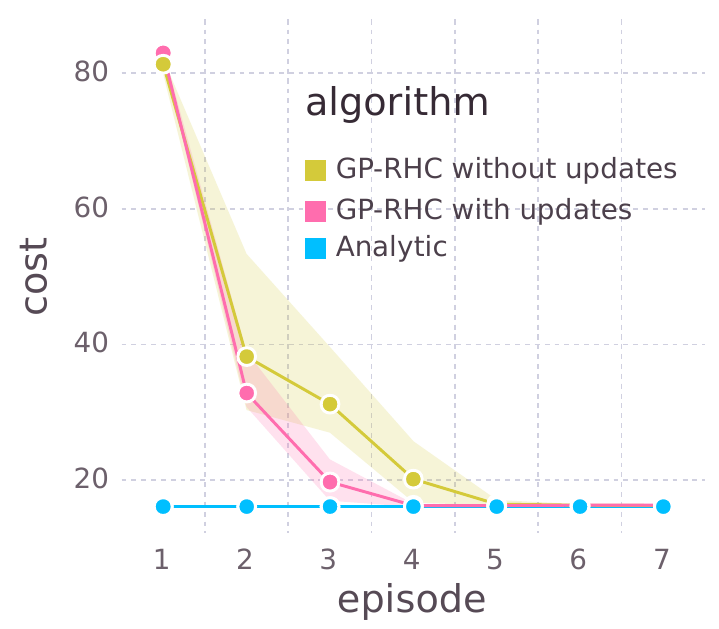}
\caption{Median cost of the constrained cartpole swing-up task (using the saturated cost function). The lower and upper bounds of the confidence envelope represents the first and third quantiles respectively. The results were aggregated over 22 runs.}
\label{fig:cartpole-costs}
\end{figure}

As seen in figures~\ref{fig:cartpole-comparison} and \ref{fig:cartpole-costs}, GP-RHC learns to solve the cartpole swing-up task in just a few episodes. In particular, by updating the dynamics model online we can achieve performance comparable to the optimal analytic model at least a full episode earlier. Another noticeable feature is the reduced variability in cost. GP-RHC, with online updates, performs more consistently and is less susceptible to variations in the initial training data. Finally, table \ref{tab:cartpoleviolate} shows that fewer constraint violations can be expected when using online updates. In fact, after the first episode only a single constraint violation was encountered, in contrast to the method without updates which incurred significantly more violations. These results indicate that GP-RHC enables fast learning and planning in safety critical conditions.

\begin{table}[h]
\caption{Percentage of experiments terminated due to constraint violations}
\label{tab:cartpoleviolate}
\begin{center}
\begin{tabular}{ccc}
\multicolumn{1}{c}{\bf EPISODE}  &\multicolumn{1}{c}{\bf UPDATES} &\multicolumn{1}{c}{\bf NO UPDATES} \\
\hline \\
1 & 100.0 & 100.0 \\
2 & 4.5 & 18.2 \\
3 & 0.0 & 13.6 \\
4 & 0.0 & 4.5 \\
5 & 0.0 & 0.0 \\
6 & 0.0 & 0.0 \\
7 & 0.0 & 0.0 \\
\end{tabular}
\end{center}
\end{table}

\subsection{AUTONOMOUS RACING}

In this section we apply GP-RHC to the autonomous racing of 1:43 scale remote control cars. \citet{liniger2015optimization} originally investigated this problem using an analytical model of the cars incorporated into a contouring control framework (see section~\ref{sec:contouring-control}). The objective is to maximize progress along the race course (depicted in figure~\ref{fig:gprace}) while remaining within the track boundaries. 

\subsubsection{BICYCLE MODEL}

Following \citet{liniger2015optimization}, the cars are modeled using a bicycle model. The cars are treated as rigid bodies and symmetry is used  to approximate the pairs of front and back tyres as single wheels. Pitch and roll dynamics are neglected so only in-plane motion is considered. In our experiments the additional complexity of tyre dynamics is ignored by assuming a no-slip model.

The state space is described by the vector $[x, y, v, \phi]$, where $x$ and $y$ denote the position of the car, $v$ the longitudinal velocity of the car, and $\phi$ the car's orientation. The control signal consists of the PWM duty cycle of the electric drive train motor and the steering angle of the front wheels. The duty cycle is constrained to the interval $[0, 1]$ and the steering angle cannot exceed 18 degrees.

\subsubsection{CONTOURING CONTROL}
\label{sec:contouring-control}

Contouring control was originally designed for industrial applications like machine tool control and laser profiling \citep{lam2010model}. The objective of the controller is to track a given reference path while maximizing some measure of progress. Often these are competing interests and we need to find a balance between speed and tracking accuracy.  In contrast to standard tracking approaches, the reference path is described only in terms of spatial coordinates. By specifying velocities and orientations the contouring controller is free to determine how the path is followed. 

Here we assume that the reference path is given by an arc length parametrized curve
\begin{equation*}
\Gamma = \{\mathbf{x} \in \mathbb{R}^q : \mathbf{x} = \gamma(s), s \in [0, l] \},
\end{equation*}
where $l$ is the total length of the path. In our experiments, the center line of the race track is used as the reference path. To find an arc length parametrization, the center line is interpolated by a cubic spline, using the method described by \citet{wang2002arc-lengthparameterized}.

Let $p_k=[x_k, y_k]$ denote the position of the car at time $t_k$. Then, the contouring error, 
\begin{equation*}
\varepsilon^c_k = \mathbf{n}(s^*_k) \cdot (p_k - \gamma(s^*_k)),
\end{equation*}
is defined as the normal deviation from the path $\gamma$, where $s^*_k$ is the value of the path parameter which minimizes the distance between the point $p_k$ and the path, and $\mathbf{n}(s^*_k)$ is the unit normal to $\gamma$ at $s^*_k$.

Calculating the contouring error requires us to determine the value of $s^*_k$ at each point along the planned trajectory. This is too computationally intensive to use as a cost function in the iterative SQP framework. To address this issue, approximations to $s^*_k$ at each point are introduced into the state. The dynamics are then augmented by the equation
\begin{equation*}
s_{k+1} = s_k + \Delta t v_k, \quad v_k \in [0, v_{\text{max}}],
\end{equation*}
where $s_k$ denotes the approximation to $s^*_k$ at time $t_k$, and $v_k$ is a virtual control input. Since the path is parameterized by arc length, $v_k$ can be thought of as the velocity of the car along the center line. For the auxiliary state $s_{k}$ to be a useful approximation we introduce a lag error term $\varepsilon^l$ defined as the distance between the points $\gamma(s^*_k)$ and $\gamma(s_k)$ along the reference path.

\begin{figure}[h]
\includegraphics[width=0.5\textwidth]{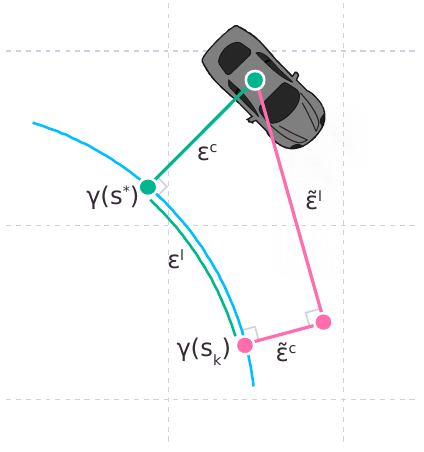}
\caption{Contouring error $\varepsilon_c$, lag error $\varepsilon_l$ and their respective approximations $\tilde{\varepsilon}_c$ and $\tilde{\varepsilon}_l$.}\label{fig:contour-error}
\end{figure}

Since neither the contouring nor lag error can be used directly in the cost function, approximations defined only in terms of $p_k$ and $s_k$ are made. The approximate contouring error $\tilde{\varepsilon}^c_k$ and approximate lag error $\tilde{\varepsilon}^l_k$ are defined as the orthogonal and tangential component of the error between the points $p_k$ and $\gamma(s_k)$,
\begin{align*}
\tilde{\varepsilon}^c_k &= \mathbf{n}(s_k) \cdot (p_k - \gamma(s_k)), \\
\tilde{\varepsilon}^l_k &= \mathbf{t}(s_k) \cdot (p_k - \gamma(s_k)),
\end{align*}
where $\mathbf{t}(s_k)$ is the unit tangent to $\gamma$ at $s_k$. It is clear from figure \ref{fig:contour-error} that $\tilde{\varepsilon}^c_k$ approaches $\varepsilon^c_k$, and  $s_k$ approaches $s^*_k$ as the lag error is reduced. Therefore, in order to get a good approximation of $s^*_k$ the lag error $\tilde{\varepsilon}^l$ is heavily penalized in the cost function \eqref{eq:racing-cost}.

Using the approximate contouring and lag errors an intermediate cost function can be defined
\begin{equation}
\label{eq:racing-cost}
\mathcal{L} = ||\tilde{\varepsilon}^l(x, y, s)||^2_{q_l} + ||\tilde{\varepsilon}^c(x, y, s)||^2_{q_c} - \alpha \Delta t v.
\end{equation}
The term $-\alpha \Delta t v$ can be thought of as a reward for progressing along the track and the weights $q_c$ and $\alpha$ represent the relative importance of fast progress and accurate path tracking. 

\subsubsection{TRACK CONSTRAINTS}

\begin{figure}[h]
\includegraphics[width=0.5\textwidth]{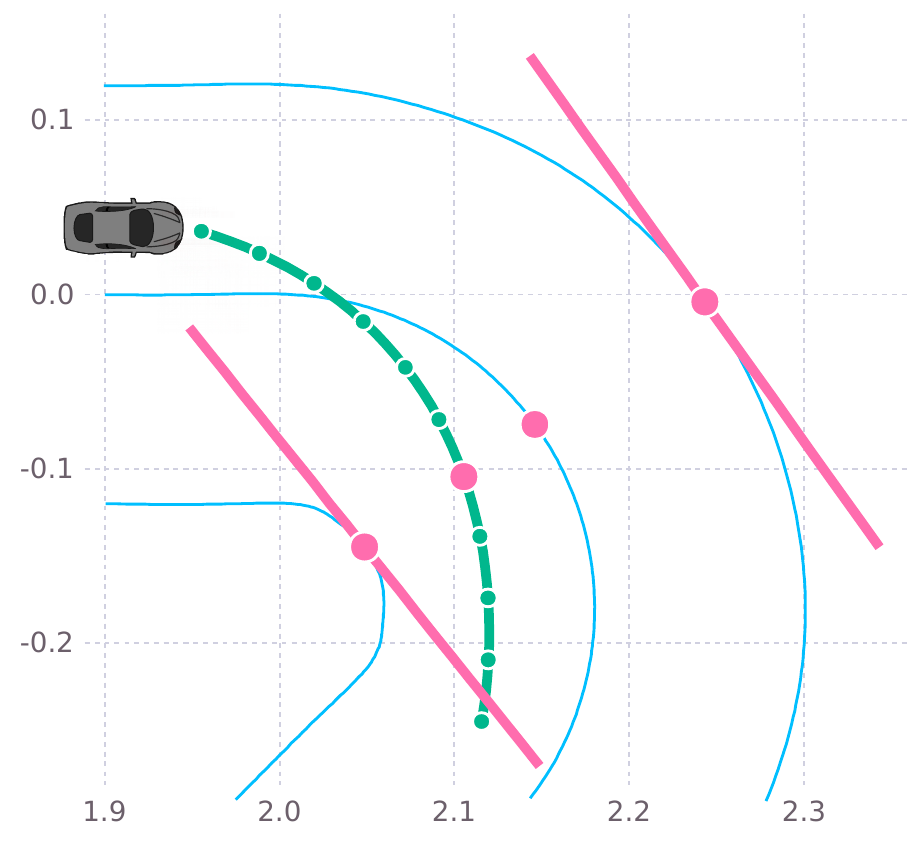}
\caption{Track constraints (pink lines) corresponding to one point (pink dot) along the planned trajectory. The central dot is the closest point lying on the center line. Tangent lines are found by projecting this point onto the track boundaries.} \label{fig:track-constraints}
\end{figure}

To ensure that the car remains within the track, limits are placed on the $x$ and $y$ components of the car's state. Each point on the planned trajectory is constrained to lie within two half spaces defined by the left and right track boundaries (see figure \ref{fig:track-constraints}). The relevant tangent lines are found by projecting the point $\gamma(s_k)$ (an approximation to the closest point on the center line) onto the track boundaries.

The planned trajectory is often along the limit of these constraints so in order to avoid infeasibility problems in practice, the constraints are softened by adding slack variables. By penalizing the slack variables heavily in the cost function the original solution of the hard constrained problem is recovered where it would admit a solution.

\subsubsection{RESULTS}

\begin{figure}[h]
\includegraphics[width=0.5\textwidth]{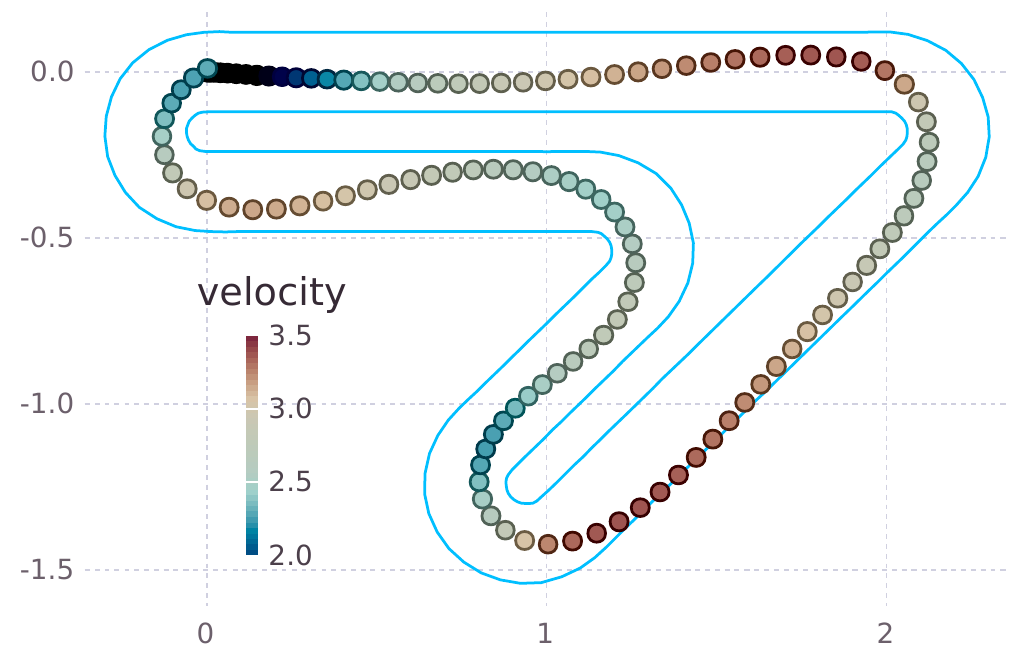}
\caption{Driven trajectory of the racing car with velocity profile given by GP-RHC. The model was initially trained on 70 points of data on a simple oval track, and was updated online during this task.}
\label{fig:gprace}
\end{figure}

Initially, 70 points of data were collected from a demonstrated trajectory around a simple oval track. 100 sample frequencies were chosen for the sparse spectrum approximation. A planning horizon of 20 time steps was used at a sampling rate of 0.03s.  We found that we could plan in real time by limiting the number of SQP iterations to 30. The race track is shown in figure \ref{fig:gprace} along with the driven trajectory. The car initially starts at rest the point $[0, 0]$ and the race is completed after one full lap. The track is 7.23m long (measured at the center line).

\begin{table}[h]
\caption{Lap Times}
\begin{center}
\begin{tabular}{ccc}
\multicolumn{1}{c}{\bf ANALYTIC}  &\multicolumn{1}{c}{\bf UPDATES} &\multicolumn{1}{c}{\bf NO UPDATES} \\
\hline \\
2.565 & 2.639 & 2.745 \\
\end{tabular}
\end{center}
\label{tab:lap-time}
\end{table}

Using a learned dynamics model, GP-RHC quickly accelerates from rest and drives trajectories that satisfy the complex track constraints (see figure~\ref{fig:gprace}). The sparse spectrum Gaussian processes are very data efficient, learning a sufficiently accurate model of the car's dynamics from just 70 data points. In particular, the training data was collected by driving around a much simpler oval track yet the learned dynamics were able to generalize well enough to effectively navigate the sharp left turn in the new race course. Table~\ref{tab:lap-time} shows the lap times of GP-RHC with and without online updates compared to the baseline analytic model. By updating the model online we improve the lap time by more than 0.1s (a relative improvement of about 3.86 percent). This represents a significant saving considering the high speeds and small length scales involved in the problem. In fact, naively scaling up the domain results in a 311m track with a lap time improvement of about 4.56 seconds. This corresponds to a 16.4km/h increase in average speed around the track when using online updates.

\begin{figure}[h]
\includegraphics[width=0.5\textwidth]{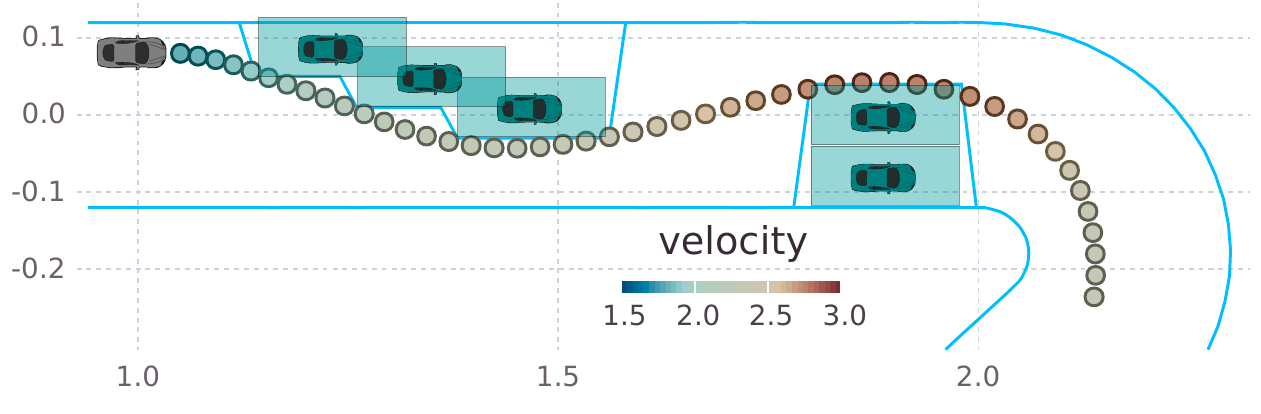}
\caption{The driven trajectory in an obstacle avoidance task. The car is able to avoid dangerous collisions by planning around the obstacles.}
\label{fig:obstacles}
\end{figure}

In addition to the racing task we also consider an obstacle avoidance problem depicted in figure~\ref{fig:obstacles}. Static obstacles, represented by the blue cars, are included by adjusting the track constraints. \citet{liniger2015optimization} determine these adjustments using a high-level planner based on dynamic programming. In our experiments, we manually specify the obstacle constraints but a high level planner could be employed in principle. Since the dynamics are independent of the constraints and cost function, GP-RHC can implicitly take advantage of all the information gained in the racing task by simply reusing the learned model with \emph{no further learning}. This could be useful in safety critical tasks where costly trials can avoided by safely learning a model of the dynamics in a simpler, safer environment.

\section{CONCLUSION AND FUTURE WORK}

In this paper we introduce GP-RHC for online learning and planning in continuous environments with non-linear constraints. This is achieved by combining receding horizon control for planning with data efficient sparse spectrum Gaussian processes for model learning.  We show that incorporating online data results in faster convergence to optimal behaviour while significantly reducing the number of constraint violations during learning --- an important feature for safety critical applications. We demonstrate our approach on a complex autonomous racing task, showing that GP-RHC enables learning from only few training points, and the ability to apply the learned model to new tasks with \emph{no additional training}. This method provides a promising approach to deploying online reinforcement learning algorithms on complex systems such as robots.

In future work plan to apply GP-RHC to real robotic systems such as quadrotors or manipulators. To achieve this goal, a possible improvement to the current method would be the ability to handle input noise. GP regression methods typically assume that the training inputs are noise free. However, in real robotic systems, sensors and filtering algorithms can introduce noise into the state estimation. This issue could be addressed, by incorporating ideas from \citet{mchutchon2011gaussian} for example, to make GP-RHC more robust.

\subsubsection*{Acknowledgements}
We thank the reviewer for their helpful insights and feedback.

\newpage

\bibliography{references}

\begin{thebibliography}{23}
\providecommand{\natexlab}[1]{#1}
\providecommand{\url}[1]{\texttt{#1}}
\expandafter\ifx\csname urlstyle\endcsname\relax
  \providecommand{\doi}[1]{doi: #1}\else
  \providecommand{\doi}{doi: \begingroup \urlstyle{rm}\Url}\fi

\bibitem[Andersson et~al.(2015)Andersson, Heintz, and
  Doherty]{andersson2015model}
Olov Andersson, Fredrik Heintz, and Patrick Doherty.
\newblock Model-based reinforcement learning in continuous environments using
  real-time constrained optimization.
\newblock In \emph{Twenty-Ninth AAAI Conference on Artificial Intelligence
  (AAAI15)}, 2015.

\bibitem[Bock and Plitt(1984)]{bock1984multiple}
Hans~Georg Bock and Karl-Josef Plitt.
\newblock A multiple shooting algorithm for direct solution of optimal control
  problems.
\newblock In \emph{Proceedings of the IFAC world congress}, 1984.

\bibitem[Boedecker et~al.(2014)Boedecker, Springenberg, W{\"u}lfing, and
  Riedmiller]{boedecker2014approximate}
Joschka Boedecker, Jost~Tobias Springenberg, Jan W{\"u}lfing, and Martin
  Riedmiller.
\newblock Approximate real-time optimal control based on sparse gaussian
  process models.
\newblock In \emph{2014 IEEE Symposium on Adaptive Dynamic Programming and
  Reinforcement Learning (ADPRL)}, pages 1--8. IEEE, 2014.

\bibitem[Carlson et~al.(2009)Carlson, Crilly, and
  Crilly]{carlson2009communication}
A.B. Carlson, P.~Crilly, and P.B. Crilly.
\newblock \emph{Communication Systems}.
\newblock McGraw-Hill Education, 2009.

\bibitem[Deisenroth and Rasmussen(2011)]{deisenroth2011pilco}
Marc Deisenroth and Carl~E Rasmussen.
\newblock Pilco: A model-based and data-efficient approach to policy search.
\newblock In \emph{Proceedings of the 28th International Conference on machine
  learning (ICML-11)}, pages 465--472, 2011.

\bibitem[Deisenroth(2010)]{deisenroth2010efficient}
Marc~Peter Deisenroth.
\newblock \emph{Efficient reinforcement learning using Gaussian processes}.
\newblock KIT Scientific Publishing, 2010.

\bibitem[Domahidi and Jerez(2014)]{FORCESPro}
Alexander Domahidi and Juan Jerez.
\newblock {FORCES Professional}.
\newblock {embotech GmbH (\nobreak{\url{http://embotech.com/FORCES-Pro}})},
  July 2014.

\bibitem[Domahidi et~al.(2012)Domahidi, Zgraggen, Zeilinger, Morari, and
  Jones]{domahidi2012efficient}
Alexander Domahidi, Aldo~U Zgraggen, Melanie~N Zeilinger, Manfred Morari, and
  Colin~N Jones.
\newblock Efficient interior point methods for multistage problems arising in
  receding horizon control.
\newblock In \emph{2012 IEEE 51st IEEE Conference on Decision and Control
  (CDC)}, pages 668--674. IEEE, 2012.

\bibitem[Gijsberts and Metta(2013)]{gijsberts2013real}
Arjan Gijsberts and Giorgio Metta.
\newblock Real-time model learning using incremental sparse spectrum gaussian
  process regression.
\newblock \emph{Neural Networks}, 41:\penalty0 59--69, 2013.

\bibitem[Kim et~al.(2004)Kim, Jordan, Sastry, and Ng]{kim2004autonomous}
H.~J. Kim, Michael~I. Jordan, Shankar Sastry, and Andrew~Y. Ng.
\newblock Autonomous helicopter flight via reinforcement learning.
\newblock In \emph{Advances in Neural Information Processing Systems 16}, pages
  799--806. MIT Press, 2004.

\bibitem[Kober et~al.(2013)Kober, Bagnell, and Peters]{kober2013reinforcement}
Jens Kober, J~Andrew Bagnell, and Jan Peters.
\newblock Reinforcement learning in robotics: A survey.
\newblock \emph{The International Journal of Robotics Research}, 2013.

\bibitem[Kouzoupis et~al.(2015)Kouzoupis, Zanelli, Peyrl, and
  Ferreau]{kouzoupis2015towards}
D~Kouzoupis, A~Zanelli, Helfried Peyrl, and Hans~Joachim Ferreau.
\newblock Towards proper assessment of qp algorithms for embedded model
  predictive control.
\newblock In \emph{Control Conference (ECC), 2015 European}, pages 2609--2616.
  IEEE, 2015.

\bibitem[Lam et~al.(2010)Lam, Manzie, and Good]{lam2010model}
Denise Lam, Chris Manzie, and Malcolm Good.
\newblock Model predictive contouring control.
\newblock In \emph{Decision and Control (CDC), 2010 49th IEEE Conference on},
  pages 6137--6142. IEEE, 2010.

\bibitem[L{\'a}zaro-Gredilla et~al.(2010)L{\'a}zaro-Gredilla,
  Qui{\~n}onero-Candela, Rasmussen, and Figueiras-Vidal]{lazaro2010sparse}
Miguel L{\'a}zaro-Gredilla, Joaquin Qui{\~n}onero-Candela, Carl~Edward
  Rasmussen, and An{\'\i}bal~R Figueiras-Vidal.
\newblock Sparse spectrum gaussian process regression.
\newblock \emph{The Journal of Machine Learning Research}, 11:\penalty0
  1865--1881, 2010.

\bibitem[Liniger et~al.(2015)Liniger, Domahidi, and
  Morari]{liniger2015optimization}
Alexander Liniger, Alexander Domahidi, and Manfred Morari.
\newblock Optimization-based autonomous racing of 1:43 scale rc cars.
\newblock \emph{Optimal Control Applications and Methods}, 36\penalty0
  (5):\penalty0 628--647, 2015.

\bibitem[Mchutchon and Rasmussen(2011)]{mchutchon2011gaussian}
Andrew Mchutchon and Carl~E. Rasmussen.
\newblock Gaussian process training with input noise.
\newblock In \emph{Advances in Neural Information Processing Systems 24}, pages
  1341--1349. Curran Associates, Inc., 2011.

\bibitem[Nocedal and Wright(2006)]{nocedal2006numerical}
J.~Nocedal and S.~Wright.
\newblock \emph{Numerical Optimization}.
\newblock Springer Series in Operations Research and Financial Engineering.
  Springer New York, 2006.

\bibitem[Pan and Theodorou(2014)]{pan2014probabilistic}
Yunpeng Pan and Evangelos Theodorou.
\newblock Probabilistic differential dynamic programming.
\newblock In \emph{Advances in Neural Information Processing Systems}, pages
  1907--1915, 2014.

\bibitem[Qui{\~n}onero-Candela and Rasmussen(2005)]{quinonero2005unifying}
Joaquin Qui{\~n}onero-Candela and Carl~Edward Rasmussen.
\newblock A unifying view of sparse approximate gaussian process regression.
\newblock \emph{Journal of Machine Learning Research}, 6\penalty0
  (Dec):\penalty0 1939--1959, 2005.

\bibitem[Rasmussen and Williams(2006)]{rasmussen2006gaussian}
Carl~Edward Rasmussen and Christopher K~I Williams.
\newblock Gaussian processes for machine learning.
\newblock 2006.

\bibitem[Rudin(2011)]{rudin2011fourier}
Walter Rudin.
\newblock \emph{Fourier analysis on groups}.
\newblock John Wiley \& Sons, 2011.

\bibitem[Vukov et~al.(2013)Vukov, Domahidi, Ferreau, Morari, and
  Diehl]{vukov2013auto}
Milan Vukov, Alexander Domahidi, Hans~Joachim Ferreau, Manfred Morari, and
  Moritz Diehl.
\newblock Auto-generated algorithms for nonlinear model predictive control on
  long and on short horizons.
\newblock In \emph{52nd IEEE Conference on Decision and Control}, pages
  5113--5118. IEEE, 2013.

\bibitem[Wang et~al.(2002)Wang, Kearney, and
  Atkinson]{wang2002arc-lengthparameterized}
Hongling Wang, Joseph Kearney, and Kendall Atkinson.
\newblock Arc-length parameterized spline curves for real-time simulation.
\newblock In \emph{In in Proc. 5th International Conference on Curves and
  Surfaces}, pages 387--396, 2002.

\end{thebibliography}

\end{document}